\documentclass[10pt,twocolumn,letterpaper]{article}

\usepackage{cvpr}              

\usepackage{graphicx}
\usepackage{amsmath}
\usepackage{amssymb}
\usepackage{booktabs}

\usepackage{array}
\usepackage{booktabs}
\usepackage{xcolor}
\usepackage{subcaption}

\usepackage[pagebackref,breaklinks,colorlinks]{hyperref}

\usepackage[capitalize]{cleveref}
\crefname{section}{Sec.}{Secs.}
\Crefname{section}{Section}{Sections}
\Crefname{table}{Table}{Tables}
\crefname{table}{Tab.}{Tabs.}

\definecolor{lightblue}{rgb}{0.54, 0.81, 0.94}
\definecolor{lightgreen}{rgb}{0.54, 0.94, 0.81}
\begin{document}

\title{Leveraging Contrastive Learning for Semantic Segmentation with Consistent Labels Across Varying Appearances}
\author{J. Montalvo, R. Alcover-Couso, P. Carballeira, A. Garcia-Martin,  J. SanMiguel, M. Escudero-Viñolo}
\maketitle

\begin{abstract}
This paper introduces a novel synthetic dataset that captures urban scenes under a variety of weather conditions, providing pixel-perfect, ground-truth-aligned images to facilitate effective feature alignment across domains. Additionally, we propose a method for domain adaptation and generalization that takes advantage of the multiple versions of each scene, enforcing feature consistency across different weather scenarios. Our experimental results demonstrate the impact of our dataset in improving performance across several alignment metrics, addressing key challenges in domain adaptation and generalization for segmentation tasks. This research also explores critical aspects of synthetic data generation, such as optimizing the balance between the volume and variability of generated images to enhance segmentation performance. Ultimately, this work sets forth a new paradigm for synthetic data generation and domain adaptation.
\end{abstract}
\section{Introduction}
A key challenge in semantic segmentation is the high human and economic costs of annotating training data, which results in limited amounts of annotated real data \cite{Cordts2016Cityscapes,AlcoverOnExploring,PCCL, montalvo2023exploiting}. To address this issue, two major research trends have emerged: Domain Adaptation (DA) and Domain Generalization (DG). These techniques leverage a source annotated domain (often synthetic images) to train models to effectively perform on a different but related target domain \cite{alcovercouso2024gradientbasedclassweightingunsupervised, alcovercouso2024layerwisemodelmergingunsupervised, hoyer2022daformer, hoyer2022hrda, hoyer2023mic}. This effective translation of knowledge is achieved by learning domain-agnostic features through methods like self-training \cite{hoyer2022daformer, Tranheden2020DACSDA}, style transfer \cite{LUO2023216,PixDa}, or feature alignment \cite{9583294, vu2019advent}. Despite impressive advancements in DA -- generally performed in an Unsupervised fashion (UDA) -- and DG for semantic segmentation, its use still underperforms compared to the ones trained with labeled target data \cite{alcovercouso2023soft, deeplabv3plus2018, NVIDIA_Semseg}.

\begin{figure}[t!]
    \centering
    \includegraphics[width=\linewidth]{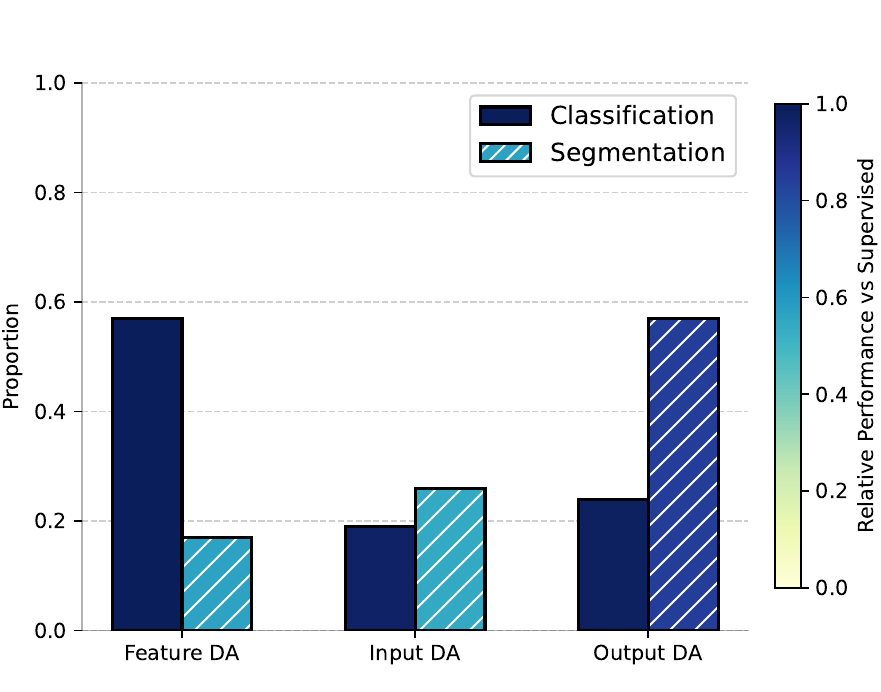}
    \caption{Visual summary of Domain Adaptation research across three tasks broadly classified into three non-exclusive categories: input domain adaptation, feature-level domain adaptation, and output domain adaptation. Around 1000 papers were scrapped and analyzed from 2018-2023.}
    \label{fig:motivation}
    \vspace{-5mm}
\end{figure}

There is theoretical room for improvement, since DA and DG achieve performance comparable to supervised methods in other tasks, such as classification \cite{Tremblay2018,hoyer2023mic,9157742}. In classification, feature-level domain adaptation techniques, 
like feature alignment, are commonly used. However, in dense prediction tasks, such as semantic segmentation, feature-level techniques are largely overlooked, in favor of output-level domain adaptation approaches, such as self-training techniques, as illustrated in Figure \ref{fig:motivation}. The primary reason for this discrepancy is that while the overall category of an image remains consistent across domains, individual pixels in semantic segmentation heavily depend on their surrounding context. Consequently, aligning features between pixels of different images can constrain the spatial learning capabilities of the model, leading to suboptimal performance.

Moreover, current UDA and DG techniques for semantic segmentation significantly underperform when used to transfer to a target scenario with harsh weather conditions \cite{hoyer2024domain,ACDC,SDV20}, limiting real-world applications like autonomous driving impractical. This underperformance can be attributed to existing synthetic urban scene datasets predominantly focusing on daylight conditions, neglecting the variability introduced by adverse weather \cite{Cordts2016Cityscapes, Richter_2016_ECCV, MVD2017}. To address the aforementioned gap, we propose a novel generation protocol that includes multiple views of each scene under different weather conditions, thereby providing semantically equivalent images to facilitate feature alignment and promote robustness across domains. Additionally, we propose a UDA and DG method to leverage the alternative versions of the semantically equivalent set of images by enforcing similar features across different weather conditions. This strategy leads to models that outperform widely-used datasets such as Synthia for DG by up to +$40\%$ mIoU, and up to +$13\%$ across four UDA methods. These benefits are extensively validated across three different alignment metrics.

This work explores three fundamental questions about synthetic dataset generation:
\begin{itemize}
    \item Q1: For synthetic data generation, what is more effective: creating a larger quantity of images, or fewer images with higher variability?
    
A1: As demonstrated in Section \ref{sec:EX}, generating fewer images with greater variability leads to better-performing models than training with larger but less diverse datasets.

    \item Q2: How can the variability of generated images be leveraged for semantic segmentation?
    
A2: In Section \ref{sec:M}, we propose a method that leverages ground-truth information to guide feature alignment between images with varying appearances. The effectiveness of this approach is validated through experiments in Section \ref{sec:EX}.
    \item Q3: Which approach better improves target domain performance: matching target domain appearance or larger variability in the source domain?
    
A3: As shown in Section \ref{sec:EX}, source domains with extensive appearance variability lead to better target performance than training with datasets tailored to match target domain appearance.
\end{itemize}

Overall, our work paves a new path for UDA and DG-based semantic segmentation through a novel synthetic data generation paradigm that enables direct application of classification-derived alignment techniques, enhancing model adaptability and performance in challenging conditions.

The rest of the paper is organized as follows: Section \ref{sec:RW} reviews the related work on synthetic data generation approaches, focusing on methods that utilize virtual environments or leverage generative deep learning models. Additionally, we outline current techniques and strategies for effective training using synthetic data. In Section \ref{sec:M}, we present our synthetic data generation protocol and introduce the proposed method to effectively leverage the generated data in UDA and DG scenarios. Section \ref{sec:EX} experimentally validates this approach, demonstrating the effectiveness of the synthetic data generation protocol in combination with the proposed training scheme for the exploitation of such data variability in UDA and DG applications. Finally, Section \ref{sec:C} closes the paper with the conclusion remarks.

\section{Related Work}
\label{sec:RW}
\subsection{Synthetic Dataset Generation}
Synthetic datasets offer several advantages over real data. They can be significantly more cost-effective than collecting and labeling large volumes of real data, especially when utilizing pre-existing data generation methods \cite{Cordts2016Cityscapes, Richter_2016_ECCV,Ros2016}. Additionally, they allow for data gathering in controlled environments and enable the collection of data for rare or dangerous events that would otherwise be difficult or hazardous to capture \cite{Alcover-Couso_2023_ICCV, MUADCH}. As deep learning models often require vast amounts of data for training, new tools for generating synthetic datasets are continually under development, ranging from fully controllable virtual environments to the more recent use of generative AI for creating synthetic data.

\paragraph{Dataset Generation Using Generative Models.}
Diffusion models \cite{ho2020denoising}, and Stable Diffusion\cite{rombach2022high} in particular, have recently gained traction in the field of generative AI due to their ability to produce high-quality images by modeling the data distribution through a series of iterative refinements. However, two major challenges persist in automatically annotating diffusion-generated images for semantic segmentation: first, accurately identifying the semantic labels present in the image, and second, spatially locating elements from each class with pixel-perfect alignment to the semantic labels. The probabilistic nature of diffusion models prioritizes plausibility and coherence over exact pixel-level accuracy, making this alignment an open research challenge. This misalignment often results in noisy labels that degrade the performance of segmentation models. While recent advances in category localization using diffusion models \cite{tang2022daam,Marcos-Manchon_2024_CVPR} have shown promise, they still struggle with misalignments in densely populated urban scenes (e.g., a car being transformed into a truck or traffic lights into traffic signs, as shown in Figure \ref{fig:missalignment}). This limitation renders diffusion models a challenging and currently suboptimal solution for urban scene UDA. To address this, we utilize virtual environments to generate reliable, pixel-level aligned ground-truth data, providing a robust foundation for urban scene DG and  UDA.  
\begin{figure}[]
    \centering
    \begin{subfigure}[b]{0.5\linewidth}
    \includegraphics[trim=5 150 450 150,clip,width=\linewidth]{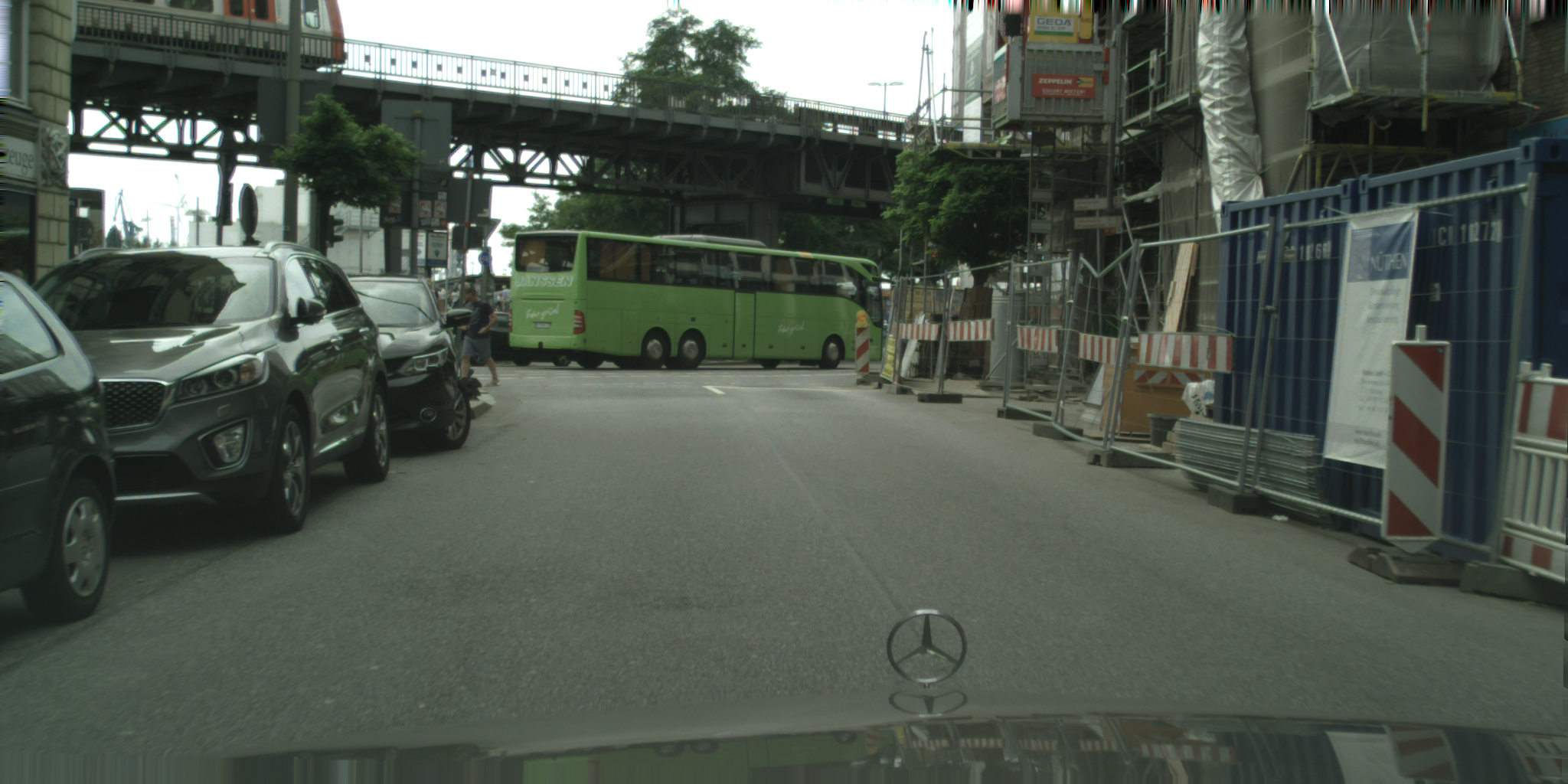}
    \caption{Original Image}
    \end{subfigure}\hfill
    \begin{subfigure}[b]{0.5\linewidth}
    \includegraphics[trim=5 150 450 150,clip,width=\linewidth]{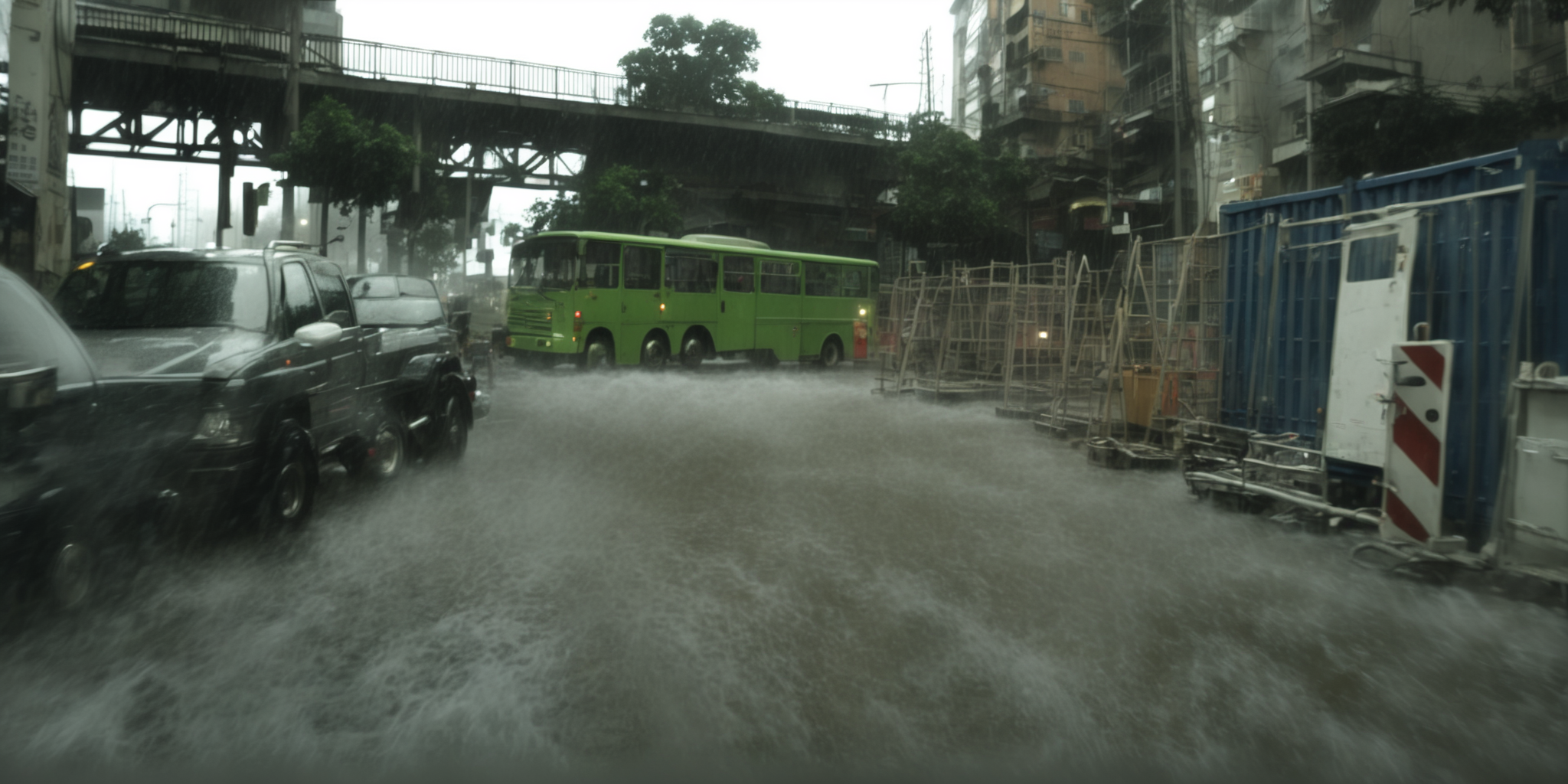}
    \caption{Style Transfered Image}
    \end{subfigure}
    \caption{Visual example of artefacts introduced by a Stable Diffusion XL model with Control Net employed for style transfer to heavy rain.}
    \label{fig:missalignment}
\end{figure}

\paragraph{Virtual Environments}

Using virtual environments to generate synthetic data provides multiple advantages over collecting real datasets. Virtual environments can simultaneously generate data and provide ground-truth annotations for various tasks, significantly reducing data collection and labeling costs. Moreover, virtual environments allow for the generation of data that would otherwise be extremely costly, dangerous, or even impossible to collect in real life, such as in-orbit satellite imagery \cite{montalvo2024spin}. 

Compared to AI generated datasets, virtual environments provide realiable and tunable data which can provide temporal and contextual relationships between frames \cite{diana2023self}. Moreover, it can provide synthetic data for very specific domains which generative models may struggle to generate. For these reasons, synthetic datasets from virtual environments are the default approximation for DG and DA synthetic data generation. 

Within virtual environments, we define two alternatives: modified game engines, such as the GTA Dataset \cite{Richter_2016_ECCV}, and synthetic engines, such as CARLA \cite{Dosovitskiy17CARLA} and LGSVL \cite{rong2020lgsvl}. Both approaches offer unlimited, fast, and cost-effective labeled synthetic data once deployed. However, the initial deployment process differs significantly. Modified game engines leverage data from video games, mapping visual instances directly to pre-existing assets, which facilitates the extraction of semantic labels. These engines often provide multiple assets or appearances for each semantic category, leading to significant intra-class variability. Despite these advantages, they face a major limitation: incorporating new semantic categories requires substantial modifications to the game’s internal engine, which reduces scalability. Additionally, some assets are inconsistently labeled; for example, in GTA V, riders' heads are often derived from pedestrian models, leading to mislabeling.

In contrast, synthetic engines require an initial investment in programming and tuning, which can be resource-intensive. However, they offer complete adaptability and scalability for future developments. Given these considerations, we build on the CARLA engine \cite{Dosovitskiy17CARLA}, which provides various urban scenarios and a diverse set of vehicles and pedestrians with customizable appearances. This adaptability enables us to tailor the engine to generate a comprehensive dataset, ideal for training and testing semantic segmentation models in urban environments.

\subsection{Usage of Synthetic Data}
Although simulators and synthetic data offer advantages such as reduced operational costs, pixel-perfect labels,  or the potential to enable research in otherwise inaccessible fields, they come with notable limitations. Synthetic environments often exhibit less variability compared to the real world due to the limited number of assets, failing to capture the almost infinite variations found in natural settings. Additionally, achieving a high degree of realism significantly escalates development costs, potentially surpassing the cost of generating real-world datasets. These discrepancies between synthetic and real-world data, known as the domain gap, are a central focus in the field of domain adaptation. Domain adaptation involves a source domain, which is a well-annotated dataset where models can be trained effectively, and a target domain, where the model is intended to be deployed. The objective is to leverage the knowledge gained from the source domain to enhance model performance in the target domain, effectively bridging the domain gap.

\paragraph{Synthetic Data Training}
In order to effectively train models with synthetic data, multiple techniques have been proposed which can be broadly classified into three non-exclusive principal categories \cite{10128983}: Input Space Domain Adaptation \cite{hoyer2022daformer, hoyer2022hrda, hoyer2023mic,Tranheden2020DACSDA,9888149}, Feature Space Domain Adaptation \cite{10.1145/3474085.3475186, wmmd, 8099590}, Output Space Domain Adaptation \cite{AK2023106172,8578878,vu2019advent, hoyer2022daformer}. Notably, as shown in Figure \ref{fig:motivation} there is a significant shift between classification and dense prediction research within domain adaptation. While classification and re-identification tasks benefit from feature space domain adaptation, dense prediction tasks predominantly utilize input and output space domain adaptation techniques like pseudo-labeling \cite{AK2023106172,8578878,vu2019advent} or image mixup \cite{Tranheden2020DACSDA, hoyer2022daformer, hoyer2023mic}.  

\paragraph{Feature Space Domain Adaptation}
The main research focus in classification and re-identification tasks is to make features indistinguishable from source and target domain, typically relying on metrics like L2 distance \cite{road, hoyer2022daformer}, Maximum Mean Discrepancy (MMD) \cite{JMLR:v13:gretton12a,Li20DCAN, wmmd} and Cosine Similarity (CS) \cite{chuah2024enhanced, wei2024self, wang2023pseudo}. Notably, these methods are tailored to learning universal features that unequivocally correlate to a specific training class. However, such training has not yet been effectively adapted to semantic segmentation, as such techniques harshly downgrade the model segmentation capabilities. We believe that as the features encode spatial and semantic information, such techniques damper the spatial information, as there are no two equivalent features semantically and spatially. Therefore, while aligning patches that encode the same semantic information is feasible \cite{hoyer2022daformer, 10128983}, those patches do not encode the same spatial information. These drawbacks set the current landscape of feature domain adaptation technique for dense predictions to focus on distillation from a teacher model using identical images, focusing on model-to-model adaptation \cite{hoyer2022daformer,hoyer2022hrda,hoyer2023mic} rather than image-to-image feature alignment, thus diverging from the trends observed in classification research (see Figure \ref{fig:motivation}). 


To overcome this drawback, we propose a protocol for generating synthetic data where multiple sets of images present drastically different visual appearances while sharing a pixel-perfect match in their ground-truths, thus allowing feature-level alignments between the image sets.

\section{Exploiting Aligned Ground-Truths with Feature Alignment }
\label{sec:M}

Here, we present our proposed framework that includes: i)  a synthetic data generation protocol that produces semantically aligned images, with multiple varied appereances, using the inherent capabilities of graphic engines, and ii) a versatile feature alignment method, that leverages the spatially aligned synthetic training data, and can be integrated into any UDA or DG framework. Additionally, our framework is designed to be compatible with various feature distance metrics, ensuring broad applicability and ease of use.
\subsection{Generating Images with Aligned Sematic Segmentation Ground-Truths}\label{generation_policy}
In this work, we explore synthetic data generation from a \textit{horizontal} perspective. Instead of creating large-scale datasets for semantic segmentation, we focus on generating diverse versions of the same data with drastically different appearances, maximizing visual variability while preserving the same ground-truth spatial information. Our goal is to modify image appearances while maintaining spatial consistency, which is crucial for effective feature alignment in semantic segmentation. We hypothesize that providing diverse appearances for all elements in the image, while fully preserving spatial structure, enhances a model’s ability to generalize to unseen data.

To achieve this, we build on the CARLA \cite{Dosovitskiy17} simulator to develop an urban scene synthetic dataset generation engine. We configure the cameras to visually resemble those used in the Cityscapes dataset \cite{Cordts2016Cityscapes}. Specifically, a camera is mounted on a vehicle driving through virtual cities. To replicate the capture conditions of Cityscapes, originally recorded from a Mercedes car, we analyzed the average height and visual characteristics of similar vehicles to determine the appropriate camera height and Field-of-View (FoV) angle. This setup ensures consistency with real-world urban scene datasets.

For generating pixel-level aligned ground truth, we capture the same scene under varying weather conditions, as well as different sun intensities and positions. These variations create visually distinct images by altering shadow direction and length, light reflection, and overall intensity. To further enhance visual diversity, we incorporate the weather categorizations from ACDC \cite{ACDC}, introducing conditions such as rain, fog, and nighttime scenarios into the dataset.

TTo achieve these modifications, we configure CARLA to ensure fully deterministic behavior using a random seed that governs all elements in the scene, including vehicle types, pedestrian models, object placements, and sensor configurations such as cameras. Using these sensors, we capture the same scene while varying weather conditions, sun intensity, and sun position. To maintain determinism and ensure dataset variability, we restart the simulation with the same seed and modify the parameters to simulate different capture conditions.

Specifically, we define four distinct capture setups:
\begin{itemize}
    \item \textbf{Noon}: A favorable condition resembling smoothly illuminated scenes where objects are clearly visible, and colors are saturated and distinct. This scenario is comparable to the setups used in synthetic datasets like Synthia \cite{Ros2016}.
    \item \textbf{Sunset}: Another favorable condition, featuring diffused lighting and slight shadows, similar to the capture conditions of the real-world Cityscapes dataset \cite{Cordts2016Cityscapes}.
    \item \textbf{Nighttime}: A harsh condition simulating low-light environments, aligned with the challenging setup defined by the real-world Dark Zurich dataset \cite{SDV20}.
    \item \textbf{Foggy}: Another harsh condition incorporating fog and clouds that obscure objects and partially occlude the scene, resembling the foggy conditions described in ACDC \cite{ACDC}.
\end{itemize}

These conditions provide a balanced representation of the diverse capture scenarios found in urban scene datasets. Figure \ref{fig:dataset_image_examples} illustrates an example of pixel-level aligned ground truth, where the same ground truth is used to generate four different RGB images under varying illumination and weather conditions.


\begin{figure}[tp]
    \centering
    \renewcommand\thesubfigure{\arabic{figure}}
    \captionsetup[subfigure]{labelformat=simple, labelsep=colon}
    \addtocounter{figure}{-2}

    \begin{minipage}[b]{0.45\linewidth}
        \centering
        \includegraphics[width=\linewidth]{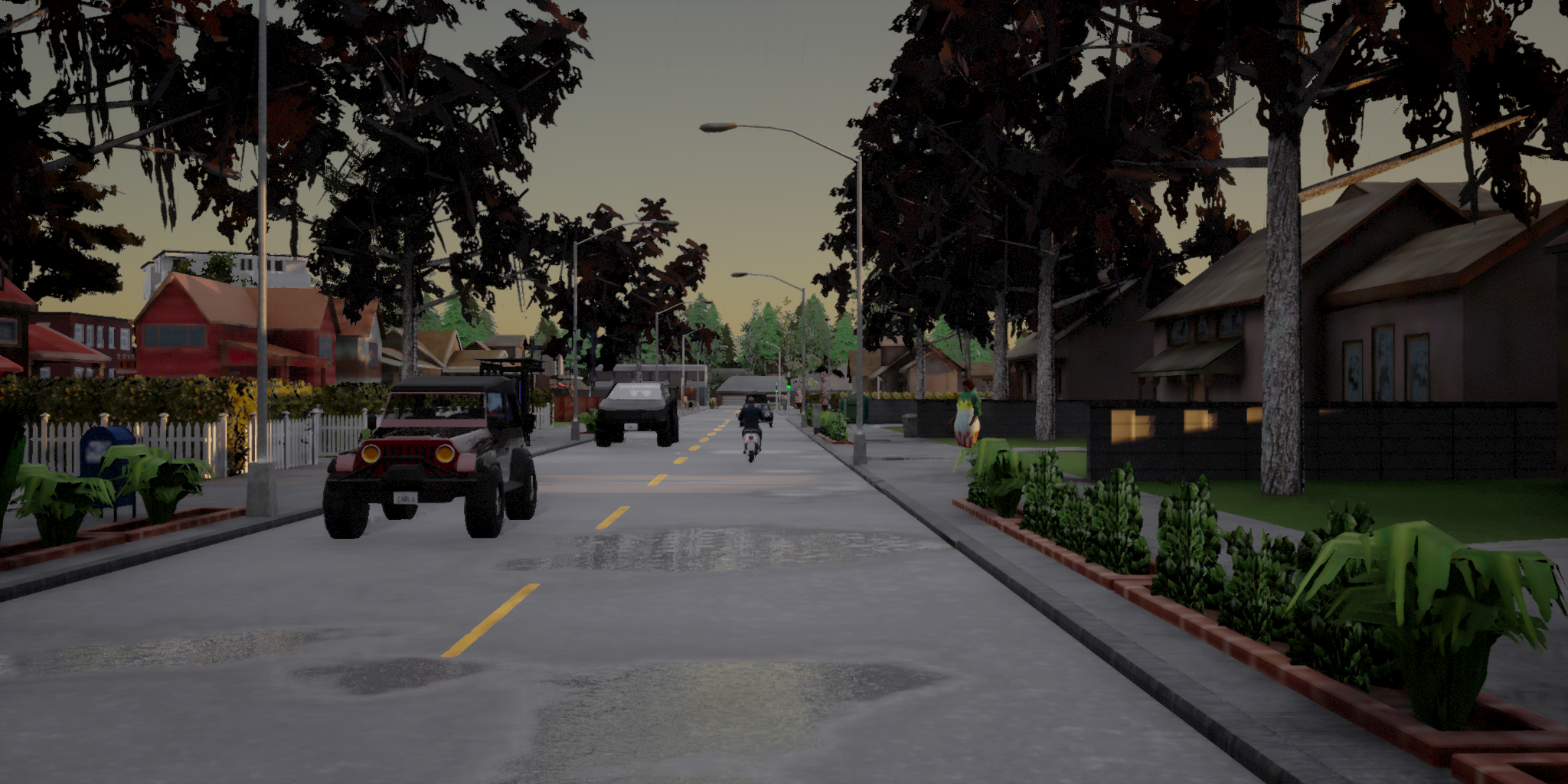}
        \subcaption{Sunset}\label{fig:1sub0}
        \addtocounter{figure}{1}
    \end{minipage}
    \hspace{0.05\linewidth}
    \begin{minipage}[b]{0.45\linewidth}
        \centering
        \includegraphics[width=\linewidth]{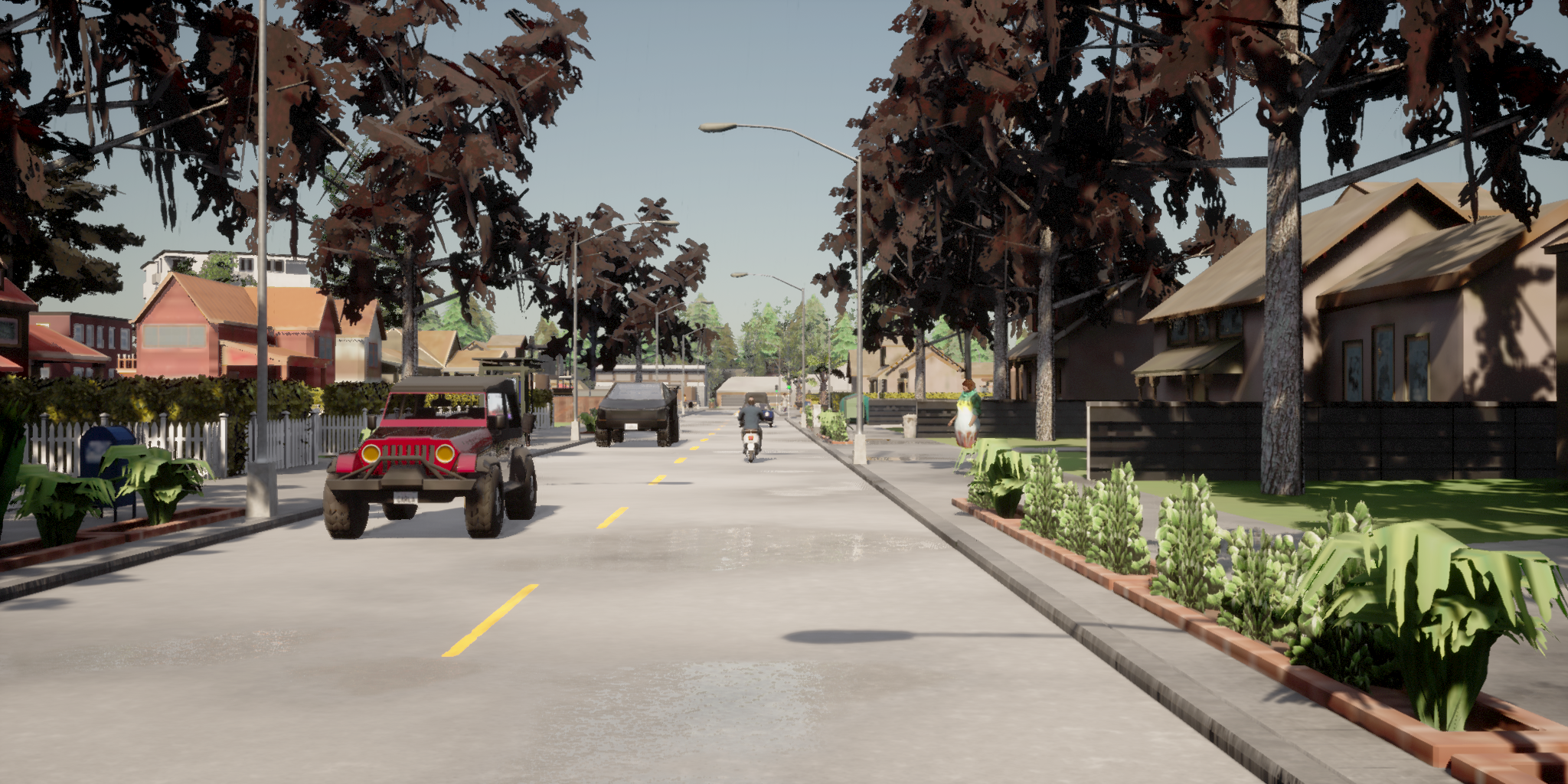}
        \subcaption{Noon}\label{fig:1sub1}
        \addtocounter{figure}{1}
    \end{minipage}
    \vspace{0.05\linewidth}
    \begin{minipage}[b]{0.45\linewidth}
        \centering
        \includegraphics[width=\linewidth]{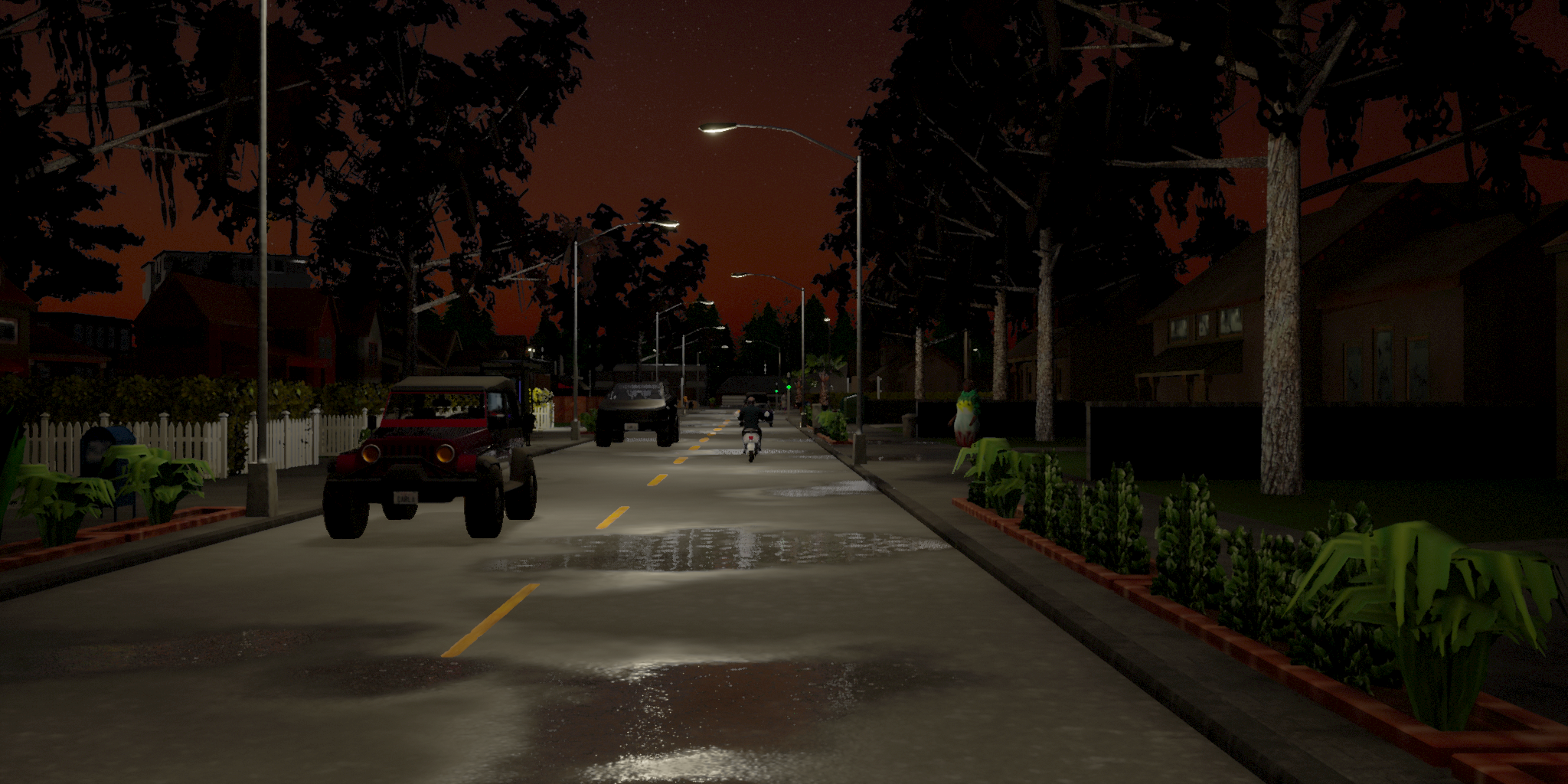}
        \subcaption{Nighttime}\label{fig:1sub2}
        \addtocounter{figure}{1}
    \end{minipage}
    \hspace{0.05\linewidth}
    \begin{minipage}[b]{0.45\linewidth}
        \centering
        \includegraphics[width=\linewidth]{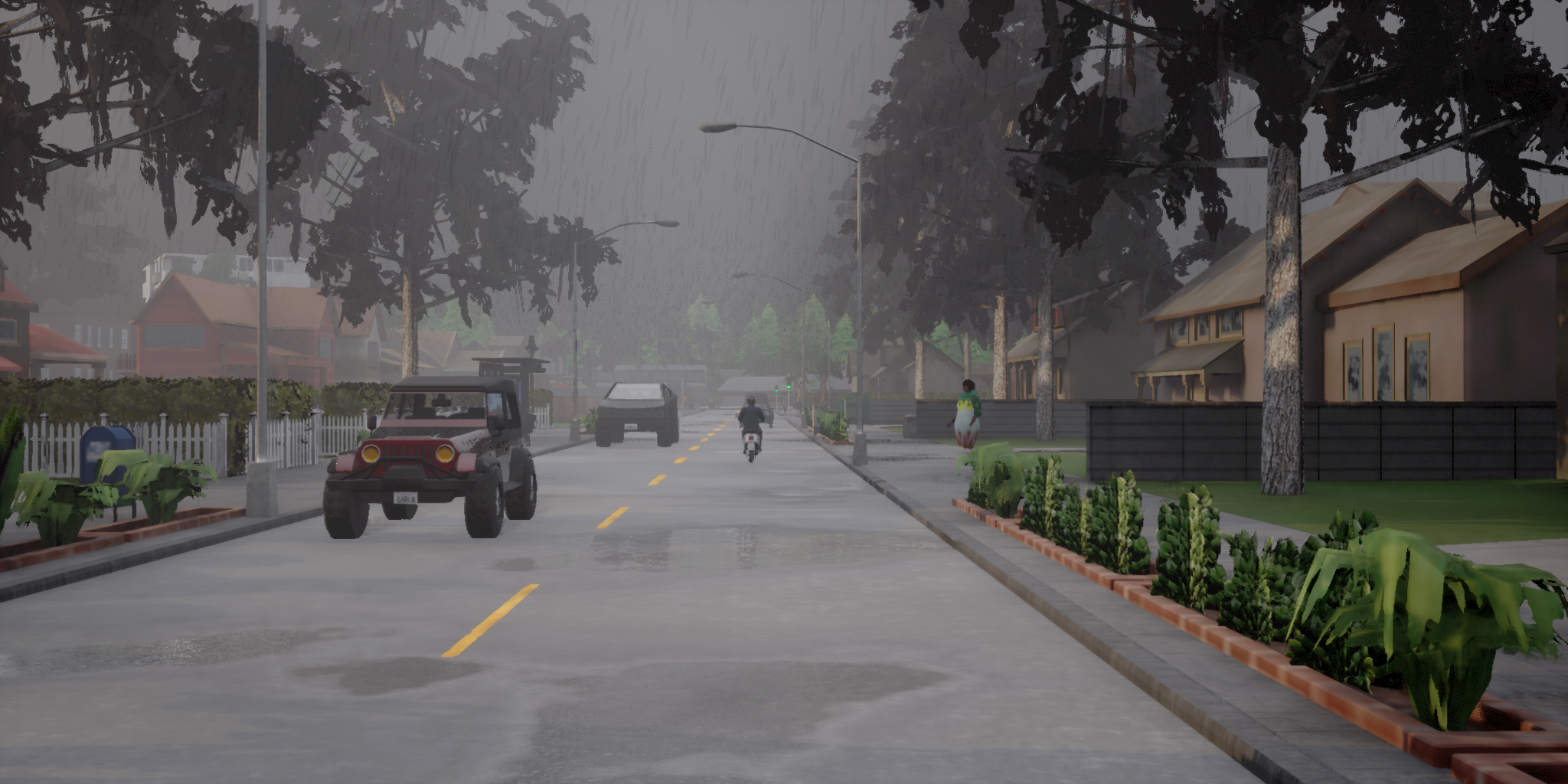}
        \subcaption{Foggy Weather}\label{fig:1sub3}
        \addtocounter{figure}{1}
    \end{minipage}
    \vspace{-0.05\linewidth}
    \begin{minipage}[b]{0.45\linewidth}
        \centering
        \includegraphics[width=\linewidth]{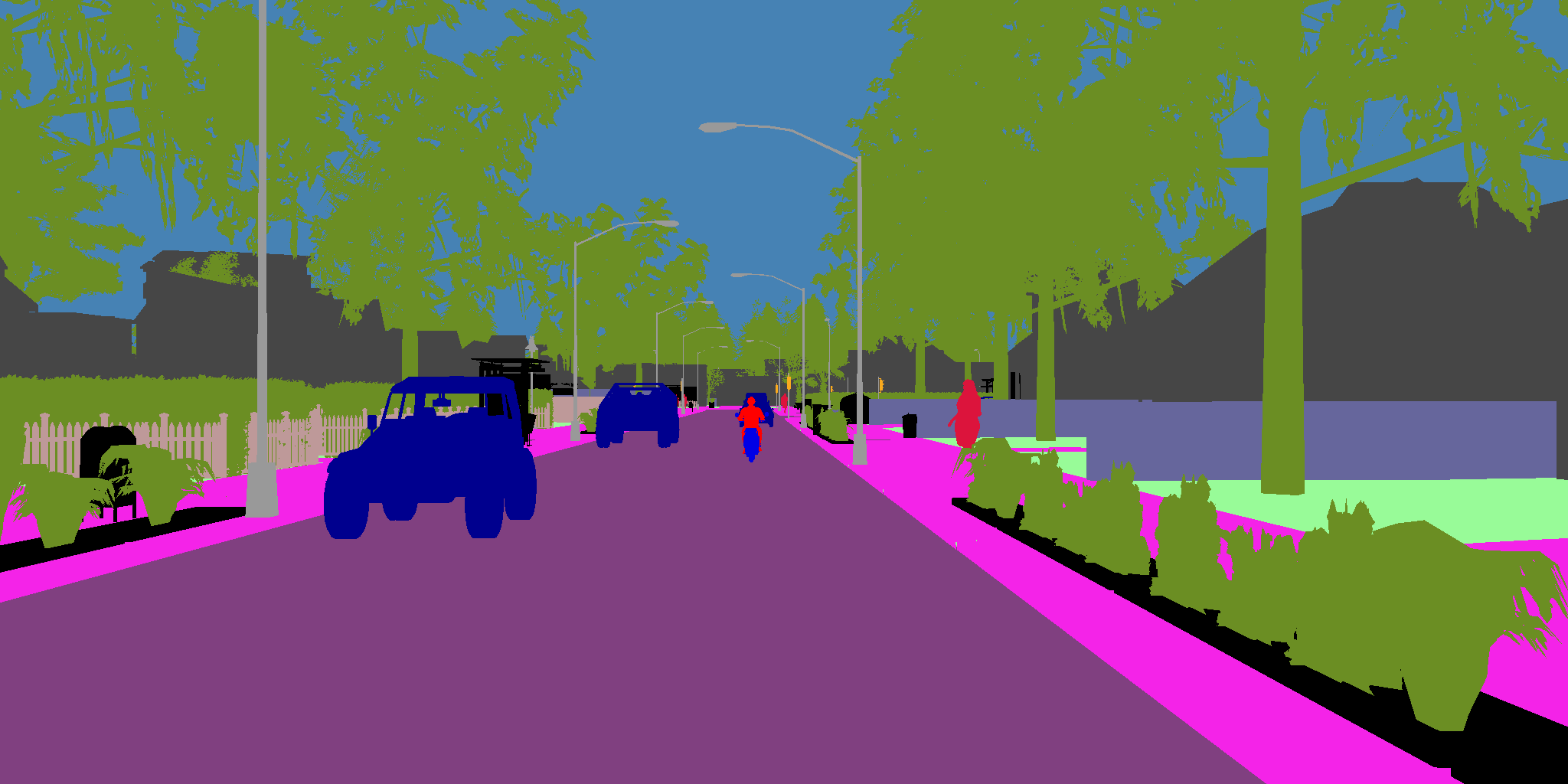}
        \subcaption{Semantic Labels}\label{fig:1subss}
    \end{minipage}
    \addtocounter{figure}{-2}
    \caption{Four different appearances for the same ground-truth.}
    \label{fig:dataset_image_examples}
\end{figure}

\subsection{Proposed Feature Alignment Module}

This subsection introduces the proposed feature alignment method within established UDA and DG frameworks that leverage different appearances of the source images. We define a straightforward module that enhances the robustness of the learned features and improves performance for: i) alternative domains in the context of DG, and ii)  a given target domain in the context of UDA. 

\begin{table*}[h]
    \centering
    \begin{tabular}{l|c l}
         Alignment ($a$) &  Formalization & Observations \\\toprule
         $L2$&  $\sum_{l=0}^L ||\mathbf{f}_l^{i,j}- \mathbf{f}_l^{i,j'}||^2$ & Euclidean distance of the features\\
         \multicolumn{1}{l|}{\raisebox{-1ex}{$MMD$}}&  $ \sup\limits_{f\in\mathcal{F}}\{\mathbb{E}f(\mathbf{f}_l^{i,j}) -\mathbb{E}f(\mathbf{f}_l^{i,j'})\}$ & \multicolumn{1}{l}{\raisebox{-1ex}{Let $\mathcal{F}$ be a class of functions f: $X \rightarrow \mathbb{R}$}}\\
         CS&  $ 1- \frac{\mathbf{f}_l^{i,j} \cdot \mathbf{f}_l^{i,j'}}{||\mathbf{f}_l^{i,j}||||\mathbf{f}_l^{i,j'}||}$ & Let each $\mathbf{f}$ be considered a vector in an euclidean space \\\bottomrule
    \end{tabular}
    \caption{Alignment loss functions ($a$)  evaluated in this work. }
    \label{tab:alignments}
\end{table*}
\paragraph{DG and UDA for Semantic Segmentation}
The goal of UDA and DG is to learn a model from a source labeled domain so it can be used without performance loss on a given unlabeled target domain (UDA), or to effectively perform on any given domain (DG). Formally, let $\mathcal{X}_S = \{{\textbf{x}_S^i, \textbf{y}_S^i}\}_{i=0}^{N_S}$  represent the labeled source domain. In the context of this work, the source domain $\mathcal{X}_S$ is our dataset. Generated using the policies described in Section~\ref{generation_policy}, each source ground-truth  $\textbf{y}_S^i$ has four alternative versions, denoted by $\textbf{x}_S^{i,j}$ with $j \in [0, 3]$ as illustrated in Figure \ref{fig:dataset_image_examples}.



\begin{figure}
    \centering
    \includegraphics[width=\linewidth]{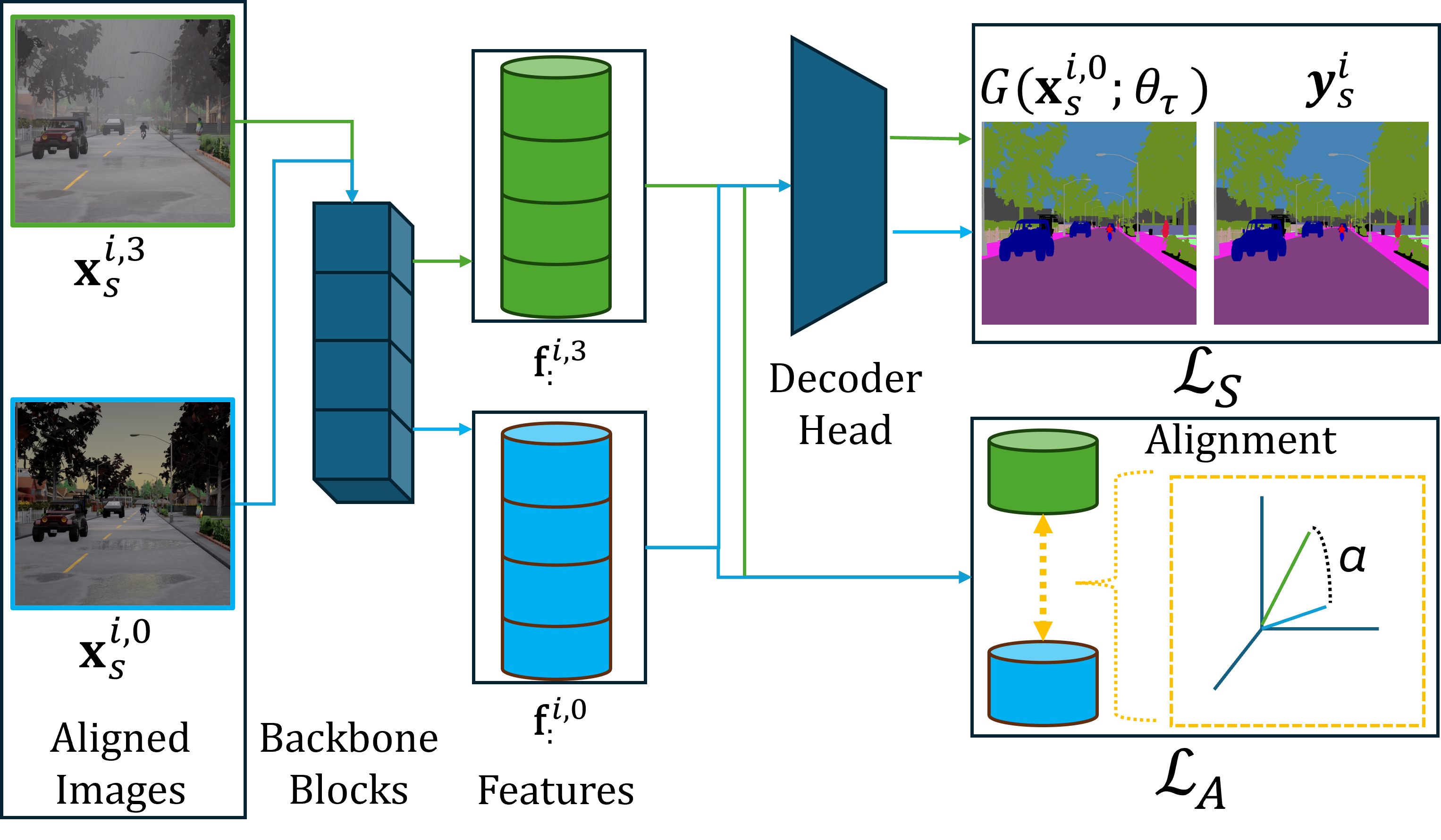}
    \caption{Visual summary of our proposed Feature Alignment Module exemplified by two visual appearances: $j=3, j'=0$.}
    \label{fig:fa}
\end{figure}

\paragraph{Feature-Level Learning with Aligned Ground-Truths}
Given that our generation protocol provides pixel-perfect, semantically aligned images, we argue that enforcing feature similarity across all layers of the model for different appearances of the same ground-truth could enable domain-agnostic training, which generalizes effectively across multiple domains. To achieve this, we propose extracting features from different levels of the model for two distinct appearances and enforcing their similarity (see Figure \ref{fig:fa}) through an alignment loss $\mathcal{L}_A$. We aim to demonstrate the versatility of our generation protocol and validate our core hypothesis: aligning features in semantic segmentation is feasible and leads to significant performance gains. To measure this feature similarity, we use three commonly used alignment functions, denoted as $a$ (see Table \ref{tab:alignments}). 

Let $G(\cdot;\theta_\tau)$ denote a model parameterized by $\theta_\tau$, where $\mathbf{f}_{s,l}^{i,j}$ represent the features extracted from layer $l\in [1,L]$ for the sample layout $i$ with appearance $j$ ($\textbf{x}_S^{i,j}$). These features are typically learned through Cross Entropy loss on the source data:
\begin{equation}
    \mathcal{L}_{S} = - \textbf{y}_S^{i} log(G(\textbf{x}_S^{i,j};\theta_\tau)).
\end{equation}

The learning objective is to ensure that features extracted  at layer $l$,  for two different image appearances $\textbf{x}_S^{i,j}, \textbf{x}_S^{i,j'} | j\neq j'$ are aligned by minimizing a feature distance $a(\cdot,\cdot)$ (see Table \ref{tab:alignments}):
\begin{equation}\label{eq:dg}
    \mathcal{L}_A = \sum_{l=0}^La(\mathbf{f}_{s,l}^{i,j}, \mathbf{f}_{s,l}^{i,j'}).
\end{equation}

We limit the analysis to two different appearances to ensure that our experimental exploration remains within the computational constraints of the state-of-the-art frameworks. The samples for the same ground-truth are sampled randomly from the four available apperances.

Summing all together, the general learning objective for the source learning is:
\begin{equation}
\begin{split}
    &\mathcal{L}_{DG} = \mathcal{L}_{S}+\lambda \mathcal{L}_A \\
\end{split}
\end{equation}
where we define $\lambda = 1/L$ as the inverse of the number of features used for the alignment, to allow scaling up to any number of features to align. 

\paragraph{Incusion of a Target Domain Unlabeled Data}
\begin{figure*}[h]
    \centering
    \includegraphics[width=.95\linewidth]{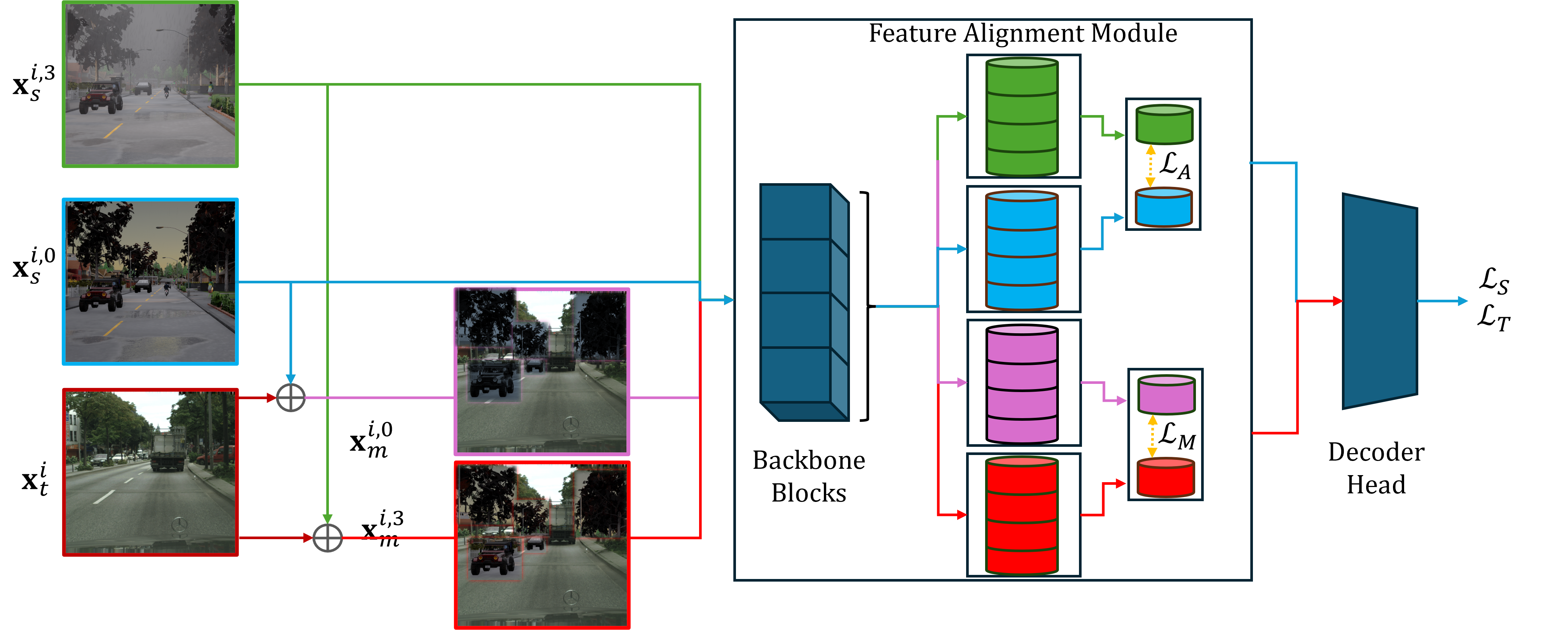}
    \caption{Visual summary of our proposed UDA training framework to exploit the different versions of the image. For visualization purpuses, image mixup is highlights overlayed source segments. For generality the target loss $\mathcal{L}_{T}$ is omited as our method is extrapolable to any target loss.  }
    \label{fig:method}
\end{figure*}
In the context of UDA, the goal is to learn a set of weights $\theta^*$ from $\mathcal{X}_S$ that can effectively generalize to a given unlabeled target domain: $\mathcal{X}_T = \{\textbf{x}_T^i\}_{i=0}^{N_T}$.


In Figure~\ref{fig:method} we show an overall schematic for our UDA method. For the target data, each UDA framework defines their own unsupervised learning objective denoted as  $\mathcal{L}_T$. Given that our framework focuses on feature alignment across different levels of the backbone, we are completely agnostic to the definition of $\mathcal{L}_T$ and is directly applicable for any $\mathcal{L}_T$. 

Nevertheless, given that UDA employs two different sets of data, we propose to further exploit the target data through image MixUp as depicted in Figure \ref{fig:method}.  Building up from the image \textbf{MixUp} technique \cite{Tranheden2020DACSDA}, we propose to generate  cross-domain images which have pixel-level aligned ground truth yielding additional data for alignment. Formally, we define the cross-domain image $\textbf{x}_m^{i,j}$ as: 
\begin{equation}
    \textbf{x}_m^{i,j} = m(\textbf{x}_S^{i,j}, \textbf{y}_S^i, \textbf{x}_T^{i'}),
\end{equation}
where $\textit{m}$ is the mixup function, which overlays the RGB cutouts of semantic instances extracted from the source domain $\textbf{y}_S^i$ on top of the target image. The overlay is defined as the element-wise  product of a binary mask $\textbf{M}^i$ and the source image:
\begin{equation}
    m(\textbf{x}_S^{i,j}, \textbf{y}_S^i, \textbf{x}_T^{i'}) = \textbf{M}^i \cdot \textbf{x}_S^{i,j} + (1- \textbf{M}^i) \cdot \textbf{x}_T^{i'},
\end{equation}
where $\textbf{M}^i$ is defined as a binary mask which filters out half of the semantic classes from $\textbf{y}s^i$. 

Intuitively, as the target regions of the mixup are identical, the generated mixed images exhibit visually distinct appearances while preserving an aligned ground truth. This setup enables feature alignment on target images, thereby enhancing the overall robustness of the extracted features.

Following the same philosophy as for the source images alignment, we employ two alternative versions of the layout $\textbf{y}_S^{i}$: $\textbf{x}_S^{i,j}, \textbf{x}_S^{i,j'} | j\neq j'$, and force their extracted features to be aligned. Formally, from these two versions of the mixed images the respective features $\mathbf{f}_{m,l}^{i,j}, \mathbf{f}_{m,l}^{i,j'}$  are used for an additional loss term employing a selected metric in Table \ref{tab:alignments}:
\begin{equation}\label{eq:layers}
    \mathcal{L}_M = \sum_{l=0}^La(\mathbf{f}_{m,l}^{i,j}, \mathbf{f}_{m,l}^{i,j'}).
\end{equation}

The complete learning objective for UDA is:
\begin{equation}
\begin{split}
    \mathcal{L}_{UDA} &= \mathcal{L}_{S}+\mathcal{L}_T+\lambda \mathcal{L}_A +\lambda \mathcal{L}_M\\
    &= \mathcal{L}_{DG} + +\mathcal{L}_T+\lambda \mathcal{L}_M,
\end{split}
\end{equation}
where $\lambda$ is defined as in DG (see Equation \ref{eq:dg}), $\lambda = 1/L$. 

\section{Experimental results}\label{sec:EX}
The goal of these experiments is to demonstrate that the proposed synthetic data generation protocol, provides data that is more effective than standard generation, both in term of yielding models with better performance on both an specific target domain (UDA) or across multiples domains (DG), while using less data. To that end, we first establish the baseline performance of each individual synthetic training subset (single appearance images). Then, we analyze the performance scalability in terms of dataset size. These initial analysis are performed using DAFormer \cite{hoyer2022daformer} as our baseline UDA framework and Hoyer, et al. \cite{hoyer2024domain} as our baseline DG framework. We conclude the experimental exploration  by validating the synthetic data  generation protocol and feature alignment across different UDA and DG methods (input, feature-level, and output domain adaptation) to assess their extrapolability. Finally, we demostrate superior performance compared to  different pixel-level aligned image generation protocols such as data augmentation and style transfer from generative models and the two most popular synthetic datasets for urban scenes ( Synthia \cite{Ros2016} and GTA \cite{Richter_2016_ECCV}). 

\subsection{Synthetic Training Dataset and Evaluation Metrics}

\subsubsection{Aligned Dataset}
As a support for our tests, we used CARLA to generate scenes on all the different scenarios from the simulator, so we also have scenery variation: from urban locations to small-town locations with more vegetation, and suburban environments. Moreover, to generate the different appearances for the same ground-truths, we choose four different sun and weather settings, offering different illumination and weather conditions: Sunset, Noon, Nighttime, and Foggy weather (see Figure \ref{fig:dataset_image_examples}). From now on, we will refer to this dataset as CARLA-Four-Aligned-Ground-Truths, or CARLA-4AGT for brevity. In total, we have generated a dataset following our proposed data generation method composed of nearly 5000 different layouts with four different appearances for each, totaling in nearly 20000 images, with semantic labels for 16 distinct classes, and the configuration files required to set up the same simulation scenes again, for future expandability. 

\textbf{Evaluation Metrics} 
For evaluation, we rely on the common metric used for semantic segmentation:  per-class intersection over union (IoU) \cite{Everingham10thepascal}, between the model predictions and the ground-truth labels. IoU measures at pixel-level the relationship between True Positives (TP), False Positives (FP), and False Negatives (FN): $IoU = \frac{TP}{TP+FP+FN}$. Additionally, we report the mean Accuracy: $Acc = \frac{TP}{TP+FN}$ across semantic categories.

\subsubsection{Exploiting the proposed Multi-Appearance Dataset}\label{sec:4.2}
We explore the benefits of using CARLA-4AGT and our Feature-Level Learning by using the DAFormer framework on the popular UDA target set Cityscapes \cite{Cordts2016Cityscapes}. \textbf{Cityscapes} is a real-image dataset with urban scenes generated by filming with a camera inside of a car while driving through different German cities. It consists of 3K images for training and 0.5K images for validation.

In all our experiments we keep the following conditions equal for consistency: models are trained for 40000 iterations, using a batch size of 2, with standard DAFormer settings.

\begin{table*}[tp]
    \setlength{\tabcolsep}{0pt}
    \centering
    \resizebox{\linewidth}{!}{%
    \begin{tabular}{l*{1}{>{\centering\arraybackslash}p{1cm}}l*{17}{>{\centering\arraybackslash}p{0.85cm}}l*{1}{>{\centering\arraybackslash}p{1cm}}r*{1}{>{\centering\arraybackslash}p{1cm}}}
        \textbf{Appearance}& \rotatebox{60}{road} & \rotatebox{60}{sidewalk} & \rotatebox{60}{building} & \rotatebox{60}{wall} & \rotatebox{60}{fence} & \rotatebox{60}{pole} & \rotatebox{60}{traffic light} & \rotatebox{60}{traffic sign} & \rotatebox{60}{vegetation} & \rotatebox{60}{sky} & \rotatebox{60}{person} & \rotatebox{60}{rider} & \rotatebox{60}{car} & \rotatebox{60}{truck} & \rotatebox{60}{motorcycle} & \rotatebox{60}{bicycle} & \rotatebox{60}{\textbf{mIoU}} & \rotatebox{60}{\textbf{mAcc}} \\ 
        \midrule
        Sunset  & 89.4 & 55.8 & 88.4 & 37.4 & 20.0 & 50.4 & 52.9 & 62.1 & 88.1 & 93.0 & 21.0 & 13.8 & 87.0 & 25.0 & 49.8 & 61.5 & 56.0 & 62.9 \\
        Noon & 89.8 & 52.4 & 87.5 & 31.7 & 17.8 & 50.1 & 52.2 & 57.9 & 86.7 &  91.0 & 32.1 & 15.3 & 89.1 & 25.2 & 51.6 & 62.7 & 55.8 & 69.4 \\
        Night & 94.8 & 65.5 & 88.0 & 43.7 & 24.7 & 46.4 & 18.3 & 15.7 & 87.7 & 93.2 & 14.1 & 12.2 & 85.1 & 22.8 & 51.7 & 61.7 & 51.6 & 67.3 \\
        Fog & 90.4 & 55.1 & 87.1 & 33.5 & 23.4 & 49.3 & 54.0 & 55.0 & 88.0 &  91.1 & 50.6 & 22.9 & 85.9 & 23.0 & 41.8 & 64.0 & 57.2 & 72.4 \\
        \midrule
        Fixed & \textbf{95.5} & \textbf{68.9} & \textbf{89.2}& 37.5 & 29.2 & \textbf{52.1} &\textbf{54.7} & 57.2 & 87.9 &  \textbf{93.6} & \textbf{41.2} & \textbf{18.6} & 84.7 & 20.5 & 41.2 & 61.1  & \textbf{58.3} & \textbf{71.9} \\
        Random & 91.5 & 55.9 & 89.1 & \textbf{43.4} & \textbf{30.1} & 50.1 & 48.4 & \textbf{59.8} & \textbf{88.3}  & 92.7 & 38.8 & 18.3 & \textbf{88.0} & \textbf{25.7} & \textbf{49.9} & \textbf{61.9} & \textbf{58.3} & 71.2 \\\midrule
        \bottomrule
   \end{tabular}}
    \caption{Per-class performance of models trained with each generated visual appearance, or a combination for DAformer on the Ours$\rightarrow$ Cityscapes UDA setup. For employing multiple visual appearances, Fixed denotes models trained seeing always for a layout $i$ two fixed appearances $x_S^{i,j}$ and $x_S^{i,j'}$. Random denotes models trained with random sampling of appearances $j$. Best performances indicated with \textbf{bold}.} 
    \label{tab:baseline_metrics}

\end{table*}

\paragraph{Impact in performance of different appearances}
In Table~\ref{tab:baseline_metrics}, we present the results after training the default DAFormer framework on different combinations of the CARLA-4AGT dataset with the goal of establishing a baseline for the performance of our dataset appearances. The table includes results for models trained on each individual visual appearance as well as combinations of appearances. We explore two training protocols for combining appearances: Fixed and Random sampling.

In the Fixed sampling protocol, a single visual appearance is randomly predefined for each layout, ensuring that the model sees only one appearance per layout throughout training. However, these predefined appearances are selected randomly to ensure a uniform distribution across the dataset. Conversely, the Random sampling protocol selects a visual appearance at random each time a layout is used for training, allowing the model to train on all possible appearances $x_S^{i,j}$, where $j \in [0,3]$.

The results showcase strong performance gains for training with all appearances across all semantic classes compared to training with a single appearance. Regarding appearances, the Foggy weather and Sunset appearances yield better models compared to Noon and Nighttime appearances. This is expected, as the target dataset (Cityscapes) is composed of cloudy images, thus having a closer visual style to the Foggy weather and Sunset appearances. As expected, the worst performance is obtained when employing Night appearances, as these are further away in style and provide harder cues of segments to the model for learning. For our next tests, we use the Random selection protocol as our baseline.

\paragraph{Measuring the performance of our feature alignment framework.}
In Table \ref{tab:segmentation_metrics}, we present results for our feature alignment framework configured with different alignment losses. In this test, we now include our feature alignment loss following the schema from Figure~\ref{fig:method}. Notably, all alignment methods surpass baseline performance (Random), underscoring the untapped potential of feature alignment in semantic segmentation\textendash a topic that may have been overlooked due to the lack of aligned data. We observe that CS achieves the highest overall performance, while MMD loss produces slightly lower results for background classes such as \textit{road} and \textit{building} but performs better for foreground classes like \textit{car} and \textit{rider}. We attribute this to the statistical nature of MMD, which accommodates greater intra-class variability, benefiting variable objects (foreground). In contrast, CS yields a more stable representation for less variable objects (background), thereby improving classification consistency for these classes.

\begin{table*}[t]
    \setlength{\tabcolsep}{0pt}
    \centering
    \resizebox{\linewidth}{!}{%
    \begin{tabular}{l*{1}{>{\centering\arraybackslash}p{1cm}}l*{17}{>{\centering\arraybackslash}p{0.85cm}}l*{1}{>{\centering\arraybackslash}p{1cm}}r*{1}{>{\centering\arraybackslash}p{1cm}}}
        \textbf{Alignment}& \rotatebox{60}{road} & \rotatebox{60}{sidewalk} & \rotatebox{60}{building} & \rotatebox{60}{wall} & \rotatebox{60}{fence} & \rotatebox{60}{pole} & \rotatebox{60}{traffic light} & \rotatebox{60}{traffic sign} & \rotatebox{60}{vegetation} &  \rotatebox{60}{sky} & \rotatebox{60}{person} & \rotatebox{60}{rider} & \rotatebox{60}{car} & \rotatebox{60}{truck} & \rotatebox{60}{motorcycle} & \rotatebox{60}{bicycle} & \rotatebox{60}{\textbf{mIoU}} & \rotatebox{60}{\textbf{mAcc}} \\ 
        \midrule
        Random & 91.5 & 55.9 & 89.1 & 43.4 & 30.1 & 50.1 & 48.4 & 59.8 & 88.3  & 92.7 & 38.8 & 18.3 & 88.0 & 25.7 & 49.9 & 61.9 & 58.3 & 71.2 \\
        \midrule
        L2  & 93.6 & 62.7 & 88.8 & 45.3 & 24.5 & 50.8 & 51.3 & 56.1 & 88.1  & 93.1 & \textbf{69.9} & 44.3 & 84.3 & 19.6  & 41.7 & 58.1  & 60.8 & 72.4 \\
        MMD & 90.1 & 53.5 & 88.2 & 41.9 & 30.5 & 49.0 & 51.3 & 55.9 & 87.8 & 92.4 & 69.8 & \textbf{44.6} & \textbf{88.4} & \textbf{28.0} & \textbf{49.9} & 57.6  & 61.2 & 71.6\\
        CS & \textbf{95.4} & \textbf{68.6} & \textbf{89.2} & \textbf{49.2} & \textbf{33.7} & \textbf{50.4} & \textbf{49.6} & \textbf{59.2} & \textbf{89.0}  & \textbf{92.5} & 68.0 & 41.1 & 86.5 & 21.7 &  48.9 & \textbf{61.6} & \textbf{62.8} & \textbf{74.8} \\\midrule
        \bottomrule
    \end{tabular}}
    \caption{Per-Class performance comparison across different alignment metrics. (Ours $\rightarrow$ Cityscapes). Best performances indicated with \textbf{bold}. Random stands for standard training with all appearances.}
    \label{tab:segmentation_metrics}
\end{table*}

\begin{figure}[t]
    \centering
    \includegraphics[width=\linewidth]{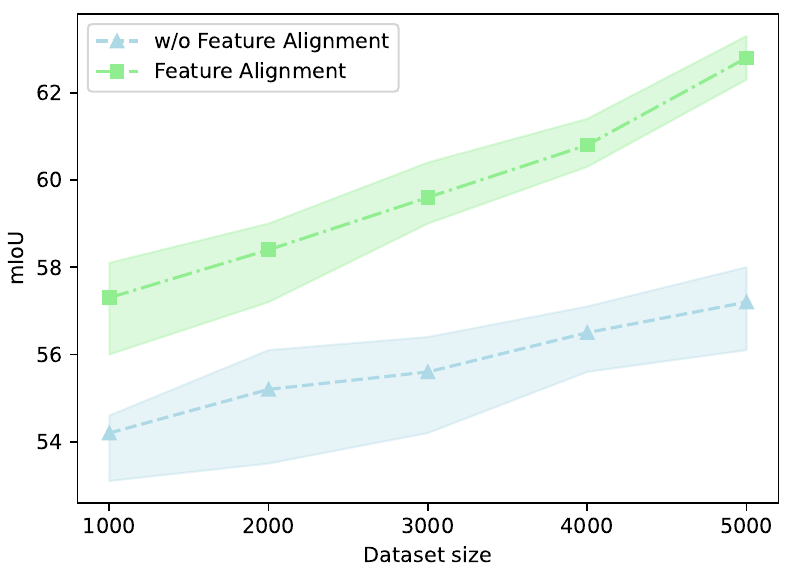}
    \caption{Performance comparison as the dataset size increases of models trained with and without feature alignment across three random seeds.}
    \label{fig:Datasetsize}
\end{figure}

\paragraph{Feature alignment allows performance scaling with available data.}
Another advantage of our framework is its improved scalability with increasing data availability compared to just generating unique ground-truth and appearance pairs. In Figure~\ref{fig:Datasetsize}, we compare the performance of our baseline method (Random) to that of the same method with feature alignment (using CS). The results illustrate how model performance evolves with a growing number of training images. Notably, the performance slope for our feature-aligned model is steeper than that of the baseline, indicating greater gains per additional data. Furthermore, with the same number of training images, our feature alignment method consistently achieves higher performance than the baseline. For instance, a model trained following the baseline (Random) strategy with 4,000 layouts  achieves 56.4\% mIoU, while using only 1,000 layouts with our feature alignment yields 57.5\% mIoU. These results underscore the effectiveness of our framework, allowing for a more efficient data generation policy. Additionally, in Figure~\ref{fig:Datasetsize} it can be observed how the employment of feature alignment allows model training to fully exploit the variability of the dataset presenting a significantly higher slope as more images are available for training.

\paragraph{Feature alignment performance scales with the number of layers employed.}
In previous experiments, we aligned features extracted the final layer of each of the residual blocks from the DAFormer architecture (see Equation \ref{eq:layers}), consisting of four residual blocks. In this analysis, we measure the performance of the final model when aligning the features of the last layer from either a single block ({\color{lightblue} lightblue}) or multiple blocks ({\color{lightgreen} lightgreen}).

\begin{figure}[t]
    \centering
    \includegraphics[width=\linewidth]{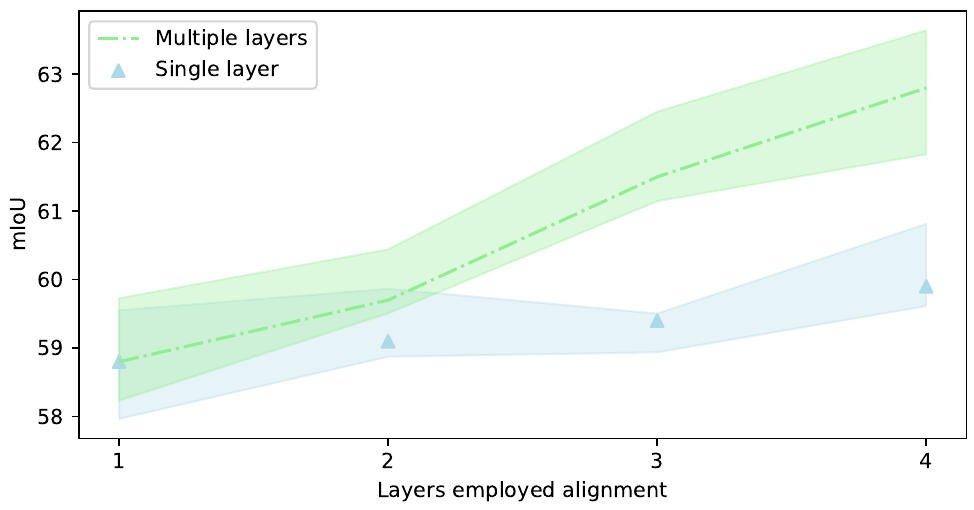}
    \caption{Performance comparison of employing features extracted from different levels of the model across three random seeds. Specifically, the {\color{lightblue} lightblue} markers denote the performance of employing a single feature vector, while the {\color{lightgreen} lightgreen} line illustrates the performance as more features are incorporated into the alignment (first, first and second, first second and third and all of them). X-axis denotes the block number, with 4 indicating last layer from last block.}
    \label{fig:layers_alignment}
\end{figure}

Figure \ref{fig:layers_alignment} shows that aligning features from later layers yields better performance, likely because the influence of these alignments trickles down to earlier layers. However, the best performance is achieved when aligning features across multiple layers, resulting in best performance, up to + 6\% mIoU increase compared to only using the first layer for alignment and up to + 8\% mIoU increase compared to not performing any alignment (see Table \ref{tab:segmentation_metrics}). 

\subsection{Exploring generalization capabilities of our framework} 
\paragraph{Datasets analyzed for comparison in Domain Generalization}
To measure the DG capabilities, we train the same DAFormer model only with source images from CARLA-4AGT and evaluate its performance on three real datasets: Cityscapes, ACDC \cite{ACDC}, and Dark Zurich \cite{SDV20}. \textbf{ACDC} is a real-image dataset composed of 2K images under adverse weather conditions (fog, night, rain, and snow). \textbf{DarkZurich} is a real-image dataset with 151 test images captured during nighttime.

\paragraph{The benefits of the proposed method transfer to different domains}

\begin{table}[]
    \setlength{\tabcolsep}{5pt}
    \centering
    \resizebox{\linewidth}{!}{%
    \begin{tabular}{lccc}
        \textbf{Appearance} & \textbf{Cityscapes} & \textbf{ACDC}& \textbf{Dark Zurich} \\
        \midrule
        Sunset & 44.9& 33.0& 18.9\\
        Noon & 44.7 & 33.0& 18.1\\
        Night & 35.8 & 26.6& 13.1\\
        Fog & 45.0& 32.9& 18.7\\
        \midrule
        Random &\textbf{46.1} &\textbf{33.2} & \textbf{20.3}\\
        \midrule
        \bottomrule
    \end{tabular}}
    \caption{Domain generalization performance comparison regarding the visual appearance of the images employed for training.  Random stands for randomly selecting an image across all four visual appearances. Best performances indicated with \textbf{bold}.}
    \label{tab:app}
\end{table}

The multi-appearance nature of CARLA-4AGT,  makes it especially suited to tackle the intricacies of multiple domains without prior access to them during training. Table \ref{tab:app} presents the domain generalization results (training only on the source domain) using the same protocol from our Random baseline. In a similar vein to the UDA results, employing multiple appearances presents better generalization results even on target datasets with low variability in appearance, such as Cityscapes and Dark Zurich, which are only comprised of daytime or night-time scenes respectively. 
 
\begin{table}[]
    \setlength{\tabcolsep}{5pt}
    \centering
    \resizebox{\linewidth}{!}{%
    \begin{tabular}{lccc}
        \textbf{Alignment} & \textbf{Cityscapes} & \textbf{ACDC}& \textbf{Dark Zurich} \\
        \midrule
        Random &46.1 &33.2 & 20.3\\
        Consistency \cite{zhao2022style} &46.7 &34.0 & 23.3\\
        \midrule
        L2 &48.5 &38.5 & 24.0\\
        MMD&48.8& 38.7 & 24.1\\
        CS &\textbf{51.0} &\textbf{39.8} & \textbf{24.9}\\
        \midrule
        \bottomrule
    \end{tabular}}
    \caption{Domain generalization performance comparison across different alignment metrics. Best performances are indicated with \textbf{bold}.}
    \label{tab:ali}
\end{table}

Our feature alignment also makes the model significantly more robust to changes in appearance, such as different cities or weather variations. Table~\ref{tab:ali} compares our Random baseline and the prediction-level consistency loss from SHADE~\cite{zhao2022style} with the results of applying the three feature alignment metrics presented in Table~\ref{tab:alignments}, all of which show substantial performance improvements over standard training approaches. Similar to our findings in UDA (Section \ref{sec:4.2}), feature alignment consistently enhances model performance compared to the baseline, regardless of the alignment metric used, once again with CS being the best-performing metric. Additionally, we show a performance increase of our method compared with the state-of-the-art domain generalization approach from SHADE \cite{zhao2022style}, which enforces logit consistency across color augmentations, whereas our method enforces homogeneous representations across all feature extraction layers. This multi-level alignment provides significant advantages across all datasets analyzed, further demonstrating the effectiveness of our approach in domain generalization.

\paragraph{The proposed dataset surpasses alternative datasets across domains}
\begin{table}[]
    \setlength{\tabcolsep}{5pt}
    \centering
    \resizebox{\linewidth}{!}{%
    \begin{tabular}{lccc}
        \textbf{Dataset} & \textbf{Cityscapes} & \textbf{ACDC}& \textbf{Dark Zurich} \\
        \midrule
        GTA & 48.9 &\textbf{39.8}& 20.4\\
        Synthia &40.5 & 31.8& 17.8\\
        \midrule
        Ours &\textbf{51.0} &\textbf{39.8} & \textbf{24.9}\\
        \midrule
        \bottomrule
    \end{tabular}}
    \caption{Domain generalization results compared to Synthia and GTA. All methods were trained with the same hyperparameters and for the same amount of iterations. Best performances are indicated with \textbf{bold}.}
    \label{tab:dg}
\end{table}

 Additionally, we compare the generalization capabilities of the model trained with CARLA-4AGT against two widely used synthetic datasets Synthia \cite{Ros2016} and GTA \cite{Richter_2016_ECCV}. \textbf{Synthia} is a synthetic urban scenes dataset for semantic segmentation, composed of 9.5K images and with 16 common semantic classes with Cityscapes. \textbf{GTA} is a synthetic dataset comprising 25K images and sharing 19 semantic classes with Cityscapes.  Notably, the GTA dataset has more intra-class visual variety than CARLA-4AGT (only for the \textit{car} class, GTAV has 150+ different car assets compared to CARLA's 18 different car assets). Table~\ref{tab:dg} shows the quantitative results of this comparison. Nevertheless, we improve performance across all target datasets, suggesting that our dataset, with our proposed feature alignment, allows the training of more generalizable models, particularly for dark scenes as illustrated by the significant performance improvements on ACDC and Dark Zurich.


\subsection{Comparisons with the state of the art}
\paragraph{The proposed framework is applicable to different UDA methods}
\begin{table*}[]
    \setlength{\tabcolsep}{0pt}
    \centering
    \resizebox{\linewidth}{!}{%
    \begin{tabular}{l*{18}{>{\centering\arraybackslash}p{0.8cm}}l*{1}{>{\centering\arraybackslash}p{1cm}}r*{1}{>{\centering\arraybackslash}p{1cm}}}
        \textbf{Method}& \rotatebox{60}{road} & \rotatebox{60}{sidewalk} & \rotatebox{60}{building} & \rotatebox{60}{wall} & \rotatebox{60}{fence} & \rotatebox{60}{pole} & \rotatebox{60}{traffic light} & \rotatebox{60}{traffic sign} & \rotatebox{60}{vegetation} &  \rotatebox{60}{sky} & \rotatebox{60}{person} & \rotatebox{60}{rider} & \rotatebox{60}{car} & \rotatebox{60}{truck} & \rotatebox{60}{motorcycle} & \rotatebox{60}{bicycle} & \rotatebox{60}{\textbf{mIoU}} & \rotatebox{60}{\textbf{mAcc}} \\ 
        \midrule
        DAFormer \cite{hoyer2022daformer} & 90.4 & 55.1 & 87.1 & 33.5 & 23.4 & 49.3 & 54.0 & 55.0 & 88.0 & 91.1 & 50.6 & 22.9 & 85.9 & 23.0 & 41.8 & 64.0 & 57.2 & 72.4 \\
        DAFormer(Ours) & \textbf{95.4} & \textbf{68.6} & \textbf{89.2} & \textbf{49.2} & \textbf{33.7} & \textbf{50.4} & 49.6 & \textbf{59.2} & \textbf{89.0}  & \textbf{92.5} & \textbf{68.0} & \textbf{41.1 }& \textbf{86.5} & 21.7 &  \textbf{48.9} & 61.6 & \textbf{62.8} & \textbf{74.8}\\\cmidrule(l){2-19} 
        HRDA \cite{hoyer2022hrda} & 94.5 & 76.0 & 90.3 & 53.2 & 24.2 & 58.4 & 62.3 & 61.4 & 87.5 & 84.9 & 52.0 & 25.3 & 86.1 & 22.8 & 63.7 & 66.3 & 63.1 & 73.7\\
        HRDA(Ours) & \textbf{96.7} & 74.2 & 89.3 & 52.1 & 19.3 & 54.7 & 61.1 & \textbf{62.5} & \textbf{88.8} & \textbf{93.0} & \textbf{76.2} & \textbf{50.8} & \textbf{88.7} & 21.3 & 61.5 & 64.3 & \textbf{65.9} & \textbf{76.3} \\\cmidrule(l){2-19} 
        
        MIC \cite{hoyer2023mic}& 93.1 & 45.7 & 90.1 & 44.3 & 19.8 & 58.0 & 64.9 & 62.1 & 88.4 & 86.8 & 77.1 & 52.0 & 88.0 & 26.9 & 56.2 & 63.4  & 63.5 & 79.0\\
        MIC(Ours) & \textbf{93.4} & \textbf{72.9} & \textbf{90.7} & \textbf{55.9} & \textbf{36.1} & 55.8 & 62.2 & \textbf{69.8} & \textbf{89.5} & 68.9 & \textbf{77.2} & \textbf{52.6} & \textbf{93.4} & \textbf{33.6} & \textbf{62.2} & \textbf{64.8} & \textbf{67.4} & \textbf{80.9}\\\cmidrule(l){2-19} 
        ADVENT \cite{vu2019advent}& 87.6 & 46.0 & 77.3 & 4.7 & 5.4 & 32.8 & 18.4 & 13.7 & 79.6 & 75.1 & 53.5 & 17.4 & 64.7 & 8.8 & 12.0 & 11.0 & 38.0 & 44.9\\
        ADVENT(Ours) & 87.4 & \textbf{46.4} & \textbf{82.7} & \textbf{16.7} & \textbf{11.4} & \textbf{37.1} & 16.1 &\textbf{24.0} & \textbf{82.7}  & \textbf{83.3} & 49.2 & \textbf{28.2} & 61.8 & \textbf{24.7} & \textbf{23.8} & 10.1 & \textbf{42.8} & \textbf{50.8}\\
        \midrule\bottomrule
    \end{tabular}}
    \caption{Per-Class performance comparison applying feature alignment with cosine similarity on other UDA methods (Ours $\rightarrow$ Cityscapes). Best performances indicated with \textbf{bold}.}
    \label{tab:comparison_other_methods}
\end{table*}
While in previous UDA experiments (Section \ref{sec:4.2}) we evaluate using DAFormer \cite{hoyer2022daformer}, to prove that our method is general and can be used with different UDA algorithms and architectures, in Table \ref{tab:comparison_other_methods} we show results after including our feature alignment for HRDA \cite{hoyer2022hrda}, MIC \cite{hoyer2023mic} and ADVENT \cite{vu2019advent}, obtaining an average performance increase (with respect to each relative baseline) of 4,3\% in mIoU across all methods.

\paragraph{The proposed simulator-based data augmentation surpasses alternatives}

Table \ref{tab:alternatives} compares the performance of models trained using data obtained from alternative methods for generating semantically aligned images. Specifically, we compare random data augmentations, such as color jittering, contrast and brightness following \cite{hoyer2024domain}, and style transfer from image-to-image diffusion models \cite{rombach2022high}. 

Notably, our simulator-based data augmentation significantly outperforms both alternatives on both UDA and DG across all target datasets. We believe that data augmentations do not introduce enough realistic variability, as shadows, reflection, and shading directions cannot be altered through image transformations. On the other hand, diffusion models tend to hallucinate new instances or modify the existing ones, resulting in misaligned labels that significantly drop the performance of the model, as it introduces noise into the learning.
\begin{table}[]
    \centering
    \resizebox{\linewidth}{!}{%
    \begin{tabular}{c|c | c c c}
         \textbf{Task}& \textbf{UDA}\cite{hoyer2022daformer} &\multicolumn{3}{c}{\textbf{DG}\cite{hoyer2024domain}}  \\\midrule\midrule
         \textbf{Method} & \textbf{Noon-C}& \textbf{C}& \textbf{ACDC}&\textbf{DZ}\\\midrule
         Data Aug& 60.2& 46.7&34.0&23.3\\
         Style Transfer&53.7&40.8&31.2&23.7 \\
         Ours& 62.8& 51.0& 39.8&24.9\\\midrule\midrule
    \end{tabular}}
    \caption{Performance comparison of models trained with different alternatives to obtain semantically aligned images. Models are trained with Cosine Similarity for alignment. Key. C: Cityscapes, DZ: Dark Zurich, Aug: Augmentation.}
    \label{tab:alternatives}
\end{table}
\subsection{Research discussion}
This paper aims to answer three research questions:
\paragraph{Q1. When generating synthetic data, is it more effective to create a larger number of images or fewer images with greater variability?}
Our experiments indicate that generating a smaller number of highly variable images results in models with improved specificity and generalization capabilities. This is evident from the results in Table \ref{tab:baseline_metrics}, where the model's specificity is tested on the Cityscapes target dataset, and in Table  \ref{tab:app}, which measures the model's generalization across different datasets. By introducing a higher degree of variability into the training data, the model becomes more adaptable to diverse visual scenarios, improving its performance not only on the target dataset (in the UDA scenario) but also in new, unseen environments (in the DG scenario).  Additionally, our framework continues to show gains even in cases where larger dataset sizes would typically yield diminishing returns (see Figure~\ref{fig:Datasetsize}). For example, with just 1K images, the introduction of controlled and known variability yields models that outperform those trained on five times more data.

\paragraph{Q2. How can the variability of generated images be leveraged for semantic segmentation?}
To address this question, we designed a training framework specifically tailored to exploit our proposed data generation protocol. Our framework leverages the pixel-level semantic alignment across different visual appearances by forcing the model to learn appearance-agnostic features at various levels. The benefits of this approach are evident not only in the model's performance (Table~\ref{tab:segmentation_metrics}), but also in the model's scalability, with performance improving as dataset size increases (see Figure \ref{fig:Datasetsize}). Models trained with our feature alignment demonstrate a consistent positive trend, while those trained without alignment show diminishing returns as dataset size grows. Additionally, Figure  \ref{fig:layers_alignment} illustrates that increasing the number of layers used for alignment further boosts performance, indicating that reinforcing alignment throughout the model enhances overall results.

\paragraph{Q3. What is more important: visual similarity to the target domain or extensive variability in the source domain?}
Our final research question addresses the trade-off between visual similarity to the target domain and extensive variability in the source domain. We evaluated this by comparing the different appearances from CARLA-4AGT that are visually tailored to the target domain with those featuring greater variability. As shown in Tables \ref{tab:baseline_metrics} and \ref{tab:app}, appearances similar to the target domain (Foggy weather and Sunset for Cityscapes, Nighttime and Foggy weather for ACDC and Dark Zurich) offer superior results in isolation compared to those with higher visual dissimilarity. However, when combining visually distinct versions, the performance surpasses that of any single dataset (Random overperforms both Fog and Sunset when evaluated in Cityscapes and Dark Zurich by up to 7\%). Additionally, when compared to training with a synthetic dataset which is more visually realistic and variable (in terms of assets employed for each category) such as GTA, a model trained on CARLA-4AGT outperforms those trained solely on GTA by leveraging CARLA-4AGT aligned ground-truths using our alignment loss (see Table \ref{tab:dg}). These results suggest that while high intra-class variability offers initial benefits, our multi-appearance framework enables learning more general features from data, even with limited intra-class variety.

\section{Conclusions}
In this paper, we introduced a novel synthetic dataset composed of pixel-level semantically aligned images across varying daytime conditions. This dataset enabled the development of a method that leverages the consistent semantics between different visual appearances, promoting a unified feature representation during the learning process. By aligning features across different images, our approach has demonstrated substantial performance improvements, both in terms of scaling with increasing data and achieving better overall results compared to similar datasets.

Our analysis further explored the optimal levels at which features should be aligned, revealing that while the final layers play a crucial role in driving performance, incorporating feature alignment across early, mid, and final layers leads to even greater performance gains. This multi-layered alignment significantly enhances the ability of the model to generalize across different visual conditions.

These findings underscore the role of thoughtful data generation in maximizing the potential of deep learning models, especially in tasks that require strong generalization across varying appearances. By carefully curating synthetic datasets with pixel-level precision and enforcing consistent feature alignment throughout the model, we can unlock improved performance and scalability in semantic segmentation tasks.
\label{sec:C}
{\small
\bibliographystyle{ieee_fullname}
\bibliography{egbib}
}

\end{document}